\documentclass{article}

\usepackage{microtype}
\usepackage{graphicx}
\usepackage{subcaption}
\usepackage{booktabs} 

\usepackage{hyperref}

\usepackage[preprint]{icml2026}

\usepackage{amsmath}
\usepackage{amssymb}
\usepackage{mathtools}
\usepackage{amsthm}

\usepackage{algorithm}
\usepackage{algorithmic}
\usepackage{multirow}
\usepackage{amsmath,amssymb,amsthm}
\usepackage[capitalize,noabbrev]{cleveref}

\theoremstyle{plain}

\theoremstyle{definition}

\theoremstyle{remark}

\usepackage[table]{xcolor} 
\usepackage[textsize=tiny]{todonotes}

\begin{document}

\twocolumn[
  \icmltitle{SJD-PV: Speculative Jacobi Decoding with Phrase Verification for \\ Autoregressive Image Generation}



  \icmlsetsymbol{equal}{*}

  \begin{icmlauthorlist}
      \icmlauthor{Zhehao Yu}{yyy}
      \icmlauthor{Baoquan Zhang}{yyy}
      \icmlauthor{Bingqi Shan}{yyy}
      \icmlauthor{Xinhao Liu}{yyy}
      \icmlauthor{Dongliang Zhou}{yyy}
      \icmlauthor{Guotao Liang}{yyy}
      \icmlauthor{Guangming Ye}{comp}
      \icmlauthor{Yunming Ye}{yyy}
    \end{icmlauthorlist}
    
    \icmlaffiliation{yyy}{Harbin Institute of Technology, Shenzhen}
    \icmlaffiliation{comp}{SIFAR}

    \icmlcorrespondingauthor{Yunming Ye}{yeyunming@hit.edu.cn}

  \icmlkeywords{Machine Learning, ICML}

  \vskip 0.3in
]



\printAffiliationsAndNotice{}  

\begin{abstract}
Speculative Jacobi decoding (SJD) is a widely used method for accelerating autoregressive (AR) image generation, but its effectiveness is often limited by token selection ambiguity. Recently, existing SJD methods mainly attempt to address this problem by relaxing verification conditions, yet the root causing such token selection ambiguity remains unclear. To figure out such reason, in this paper, we conduct a detailed analysis, and then find that image semantics are often encoded across multiple consecutive tokens, while current methods verify tokens individually, which breaks semantic continuity and amplifies token ambiguity.
To this end, instead of performing the speculative verification on token level, we turn into the token-phrase level, and then present a novel speculative Jacobi decoding with phase verification for accelerating AR image generation. Specifically, we first constructs a token phrase library from large-scale image datasets to capture token phase that represent meaningful semantic. Then, we treat the token phrase library as a prior and design a token-phrase-level verification strategy to perform parallel speculative verification. In particular, our method is plug-and-play and can seamless integration with existing SJD methods. Extensive experiments on various datasets show our method can achieve significant acceleration after applying it into existing methods.
\end{abstract}

\section{Introduction}
\label{sec:intro}

\begin{figure}[t]
  \centering
  \includegraphics[width=\linewidth]{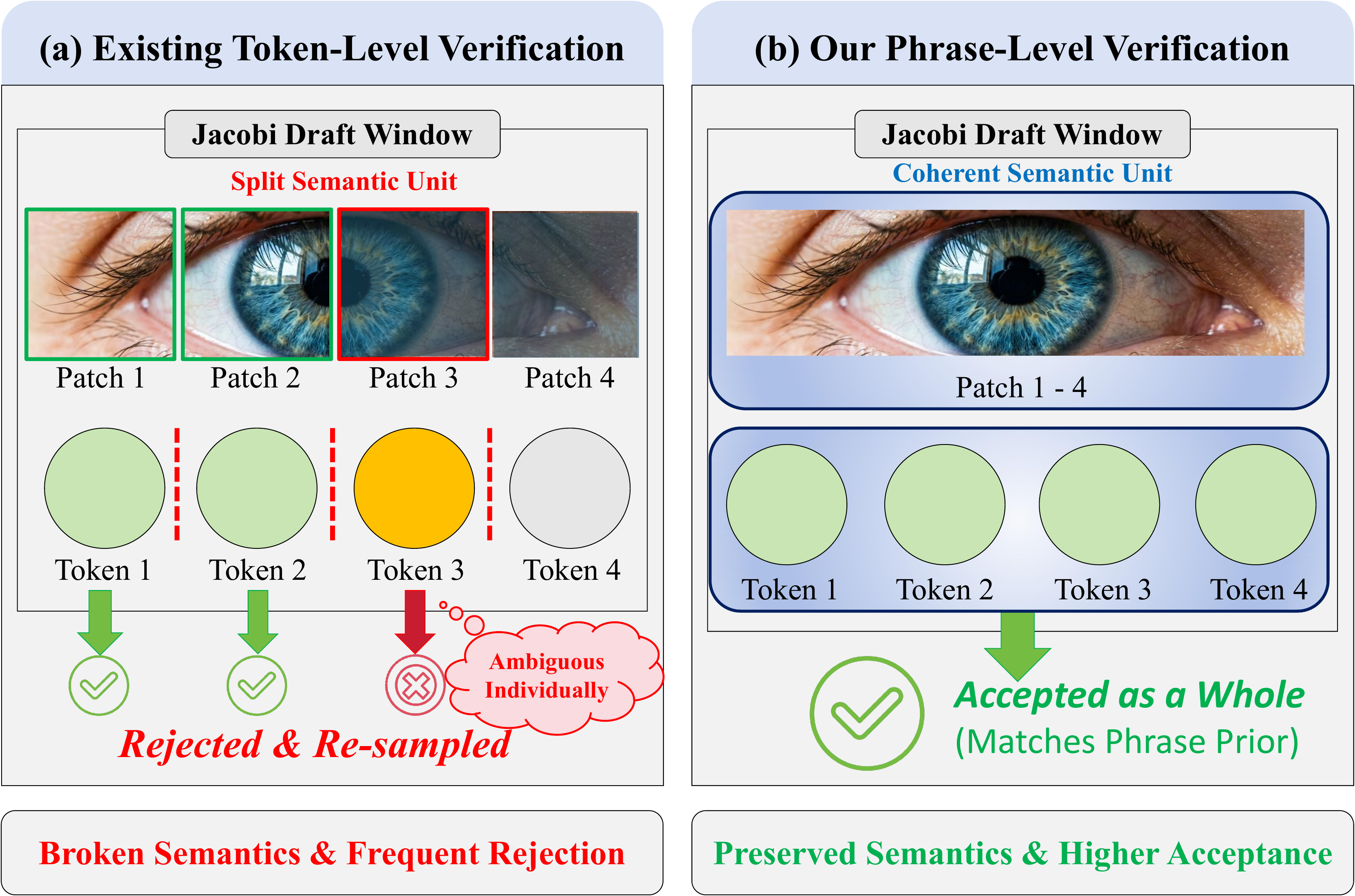}
  \caption{(a) \textit{Existing Token-Level Verification} (e.g., SJD): Tokens are verified individually, often causing rejection of locally ambiguous tokens despite their contextual coherence, which breaks visual semantics.(b) \textit{Our Phrase-Level Verification (SJD-PV)}:Contiguous tokens are grouped against a pre-constructed phrase library and verified jointly as coherent phrases  This preserves coherent visual semantics to resolve local ambiguity, improving acceptance and accelerating AR image generation.}
  \label{fig:introduction}
\vspace{-2mm}
\end{figure}
Autoregressive (AR) models have become a dominant paradigm in image generation, offering fine-grained control and exceptional visual fidelity by modeling pixel or patch-level dependencies sequentially~\cite{weng2025scaling,liu2024lumina}. Despite their impressive expressiveness, AR models suffer from inherent inefficiency during inference, which is because each token must be generated sequentially, leading to high latency and limited scalability in large-scale visual synthesis. To address the issue, Speculative Jacobi Decoding (SJD) \cite{SJD} is proposed, which aims to perform Jacobi-style iterative speculative and verification for accelerating AR image generation. Although SJD method has shown superior performance, but its effectiveness still is limited by token selection ambiguity, where AR models assign uniformly low probabilities to tokens, which significantly reduces the acceptance rate of SJD, hampering the performance of speculative decoding. 

Recently, some studies have attempted to address token selection ambiguity from a relaxed verification perspective, i.e., loosening the strict one-to-one matching constraint between speculative and verification tokens. For example, LANTERN~\cite{LANTERN} introduces a relaxed acceptance condition leveraging token interchangeability in the latent space, while Grouped Speculative Decoding (GSD)~\cite{GSD} verifies clusters of semantically consistent tokens. Although such relaxation effectively reduces false rejections and accelerates autoregressive image generation by broadening the acceptance scope, the underlying cause of token selection ambiguity remains largely unexplored.


To investigate the root cause, we conduct a detailed analysis of visual token sequence and observe a critical phenomenon: image semantics are not isolated to a individual token but are inherently encoded as stable, recurring patterns across multiple consecutive tokens. However, as shown in Figure~\ref{fig:introduction}(a), existing methods verify tokens strictly individually. This structural constraint forces coherent semantic units to be split, which breaks the integrity of semantic information, fragments token probabilities, and increases local uncertainty, thereby significantly exacerbating token selection ambiguity (see Section~\ref{sec:motivation} for more details). 
To this end, as shown in Figure~\ref{fig:introduction}(b), we propose a novel method named \textbf{Speculative Jacobi Decoding with Phrase Verification (SJD-PV)}. Different from existing methods, SJD-PV performs the parallel speculative verification step at the token-phrase level rather than on individual tokens. This design eliminates the structural constraint that forces semantic splitting, thereby effectively preserves visual semantic integrity and aligns with the natural structure of visual semantics. As a result, SJD-PV achieves higher speculative acceptance and better image generation quality for AR image generation.
Specifically, SJD-PV is built upon two key steps: \textbf{Phrase Library Construction} and \textbf{Phrase-Level Verification}. In the first step, we construct an offline token \textbf{Phrase Library} by statistically analyzing large-scale image datasets and extracting contiguous token sequences that often appear together. These sequences constitute explicit semantic priors representing coherent semantic units. In the second step, we leverage these semantic priors to perform \textbf{Phrase-Level Verification}. During this verification, whenever a draft token sequence matches a library entry, we identify it as a candidate phrase and do not verify the tokens within the phrase individually. Instead, we evaluate the joint probability of the entire candidate phrase. If the phrase is validated, all tokens within the phrase are treated as a coherent unit and accepted in a single step. Through this strategy, SJD-PV preserves coherent visual
semantics to resolve local ambiguity and consequently boosts the acceptance rate of draft tokens, thereby accelerating the decoding process.

Our main contributions can be summarized as follows:
\begin{itemize}
    \item We conduct a detailed analysis of visual token sequences and reveal a key insight: visual semantics are inherently encoded across multiple consecutive tokens rather than isolated positions. This observation suggests that the verification granularity should be aligned with these coherent semantic units (phrases) instead of a single token, motivating us to elevate the verification process from the token level to the phrase level.
    \item We introduce \textbf{SJD-PV}, a training-free, plug-and-play framework that performing the speculative verifacation on token-phrase level. By constructing a phrase library as a statistical prior, SJD-PV preserves the integrity of visual semantics during verification. This design effectively resolves the local ambiguity in existing methods, significantly improving the acceptance and accelerating AR image generation.
    \item We conduct extensive experiments on standard image generation benchmarks. Both quantitative and qualitative evaluations show that our SJD-PV can achieve significant acceleration without compromising visual quality.
    Furthermore, by integrating our method with existing SJD variants, we further improve their performance. This demonstrates the effectiveness of SJD-PV.
\end{itemize}

\section{Related Work}
\label{sec:formatting}

\subsection{Autoregressive Image Generation}

In recent years, AR image generation models have shown strong performance in fine-grained dependency modeling and precise controllability~\cite{gpt4,touvron2023llama,agarwal2025cosmos,weng2025scaling}. At present, existing AR approaches can be broadly categorized into three families:
\textbf{(1) Pixel-based AR.} These models aims to generate image directly in pixel space by predicting the next pixel (or channel) in raster-scan order, as in PixelRNN~\cite{van2016pixel}, PixelCNN~\cite{van2016conditional}, and iGPT~\cite{chen2020generative}. 
\textbf{(2) Token-based AR}, This types of models aims to generate image in discrete token space by first converting images into sequences of discrete tokens via learned codebooks (e.g., VQ-VAE~\cite{van2017neural}, VQ-GAN~\cite{esser2021taming}) and then perform autoregressive modeling in a next token prediction manner, such as DALL\@.E~\cite{ramesh2021zero} and Lumina-mgpt~\cite{liu2024lumina}. 
\textbf{(3) Scale-based AR.} These line of works follow a coarse-to-fine paradigm, representing an image as a hierarchy of token maps and autoregressively predicting the next scale. By generating entire token grids at progressively higher resolutions, they allow within-scale parallelization and support strong high-resolution synthesis, as demonstrated in recent studies such as~\cite{tian2024visual, lee2022autoregressive}. 
Despite their strong controllability and synthesis quality, AR decoders inherently operate in a strictly sequential manner. As a result, image synthesis often involves thousands of sequential decoding steps, and this strict token-by-token dependency becomes a major source of inference inefficiency.

In this work, we mainly focus on accelerating autoregressive image generation, especially on Token-based AR approaches and present a novel speculative Jacobi decoding framework from the perspective of visual phase to achieve AR generation acceleration, i.e., performing speculative in token level but parallel verification in the phase level. 

\subsection{Accelerating Autoregressive Image Decoding}
Accelerating autoregressive image decoding aims to reduce this token-by-token latency of autoregressive image generation while preserving image quality. Existing methods can be broadly categorized into two types:
\textbf{(1) Training-based approaches.} This types of  methods \cite{chen2020generative, wang2025parallelized, song2021accelerating, esser2024magvit2} aim to accelerate autoregressive decoding by transforming the inherently sequential token generation into a partially parallel forms, where multiple tokens can be generated simultaneously. For example, in \cite{chen2020generative}, Chen et al. propose a collaborative decoding strategy that coordinates large and small models across scales to accelerate inference. In  \cite{wang2025parallelized}, Wang et al. propose a parallel generation strategy that leverages token dependency patterns, generating weakly dependent tokens in parallel while maintaining sequential decoding for strongly dependent ones. \textbf{(2) Training-free approaches.} These methods, without training new components or altering the base model's distribution, reduce required evaluation iterations and idle time by adopting more efficient vertification, acceptance rules or system-level scheduling. For example in ~\cite{leviathan2023fast}, Leviathan et al. introduce a small drafter proposes multi-token drafts, verified in parallel by the base model. In ~\cite{SJD}, Teng et al. image decoding is cast as speculative Jacobi updates, accepting phrases per step without retraining.Jang et al.~\cite{LANTERN} relaxed acceptance for visual AR exploits token interchangeability to commit longer spans. 
Training-free approaches preserve the base sampling distribution, incur no retraining cost, and are plug-and-play across models and datasets. Therefore, in this work, we focuses on the training-free paradigm: we introduce a \emph{phrase-level verification} rule that seamlessly integrates with Speculative Jacobi Decoding (SJD) \citep{SJD}, thereby reducing latency and function evaluations without sacrificing visual quality.
\section{Preliminaries}
\label{sec:prelim}
\subsection{Problem Definition}
\label{subsec:problem}
Autoregressive image generation models synthesize images by sequentially predicting discrete visual tokens conditioned on previously generated ones. 
Formally, given an image represented as a sequence of tokens $\mathbf{x} = (x_1, x_2, \ldots, x_N)$ from a codebook $\mathcal{X}$, an AR model parameterized by $\theta$ defines the joint distribution as:
\begin{equation}
    p_{\theta}(\mathbf{x}) = \prod_{i=1}^{N} p_{\theta}(x_i \mid x_{<i}),
\end{equation}
where $x_{<i}$ denotes all preceding tokens. 
During inference, each token is generated through an iterative sampling process. This sequential dependency severely limits the generation efficiency, especially for high-resolution images where $N$ is large. In this paper, our goal is accelerating autoregressive image generation, i.e., reducing the decoding latency by generating and verifying multiple tokens in parallel, while ensuring that the resulting images remain visually coherent to the input condition.

\subsection{Speculative Jacobi Decoding (SJD)}
\label{subsec:sjd}
A representative acceleration method is Speculative Jacobi Decoding (SJD~\cite{SJD}). Its key idea is to parallelly draft multiple future tokens within a fixed window and then verify them in sequence using a single forward pass of the target model. Specifically, in each Jacobi iteration, SJD drafts a parallel set of candidate tokens conditioned on the current prefix and performs a \emph{single} forward pass through the target model to compute advanced conditionals for the entire window. Then, a stochastic accept–resample test is applied sequentially from left to right to verify and commit as many tokens as possible before sliding the window forward. This iterative schedule maintains SJD’s low computational cost (one target evaluation per window) while effectively reducing the number of function evaluations (NFE) by allowing multiple tokens to be accepted in a single iteration.

\section{Methodology}
\subsection{Motivation Analysis.}
\label{sec:motivation}

Although SJD methods have demonstrated superior performance, recent studies suggest their effectiveness remains constrained by token selection ambiguity~\cite{GSD, LANTERN}, where AR models often assign uniformly low probabilities to tokens. However, the root cause of this ambiguity—specifically, why certain tokens consistently exhibit low acceptance during Jacobi iterative speculation—remains unexplored. To figure out the reason, in this section, we first conduct an in-depth analysis on visual tokens, image semantics, and their corresponding relation.


\begin{figure}[t]
  \centering
  \includegraphics[width=\linewidth]{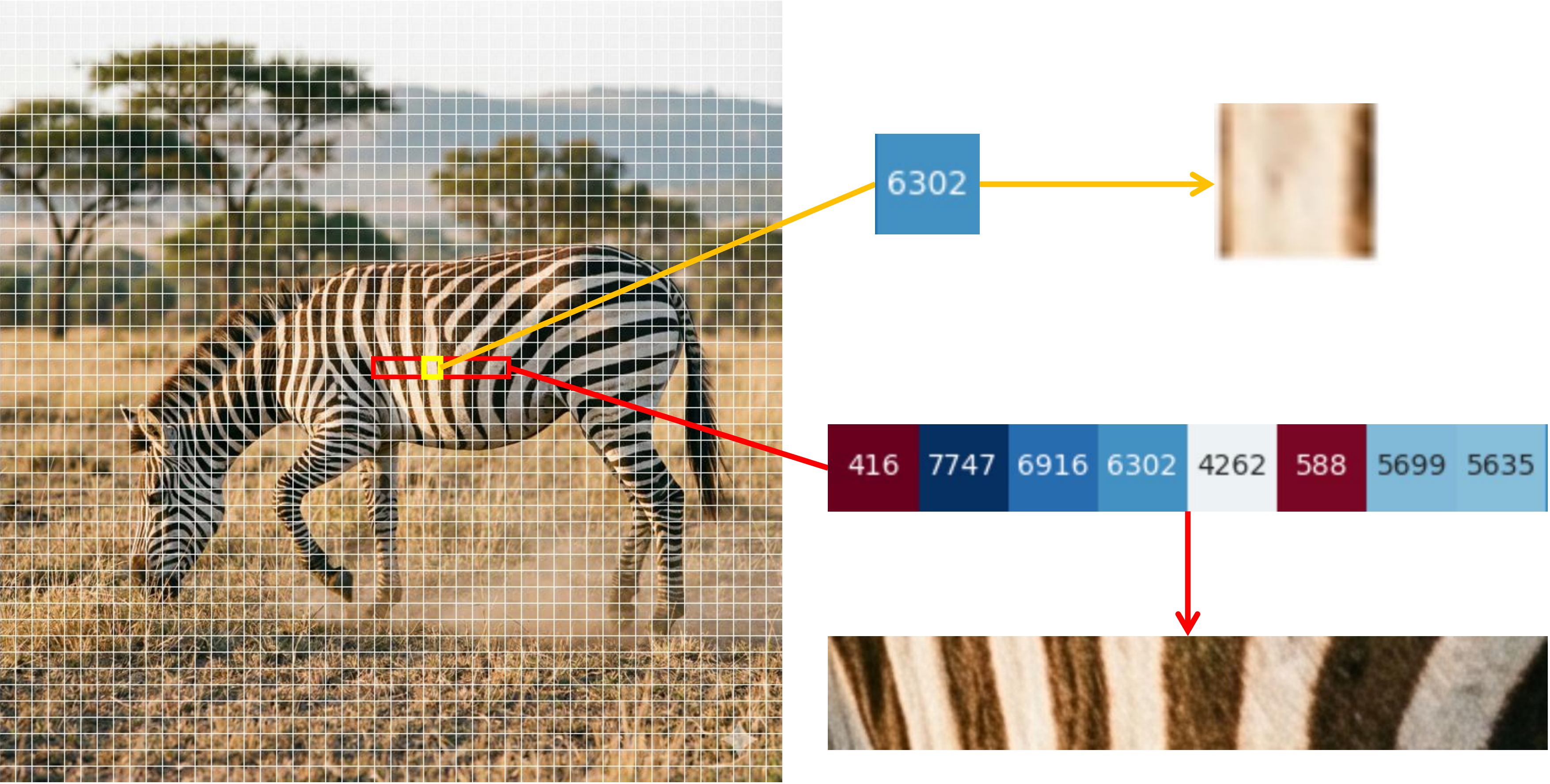}
  \caption{A toy example of the token phrase's visual semantic. While the isolated patch for Token \#6302 (top) presents an ambiguous texture, placing it within its contiguous sequence (bottom) exhibits a distinct zebra stripe structure. This validates that visual semantics are fundamentally defined by the collective behavior of neighboring tokens rather than stored independently.}
  \label{fig:visual_compare}
\end{figure}

Specifically, considering that visual semantics naturally have strong spatial locality~\cite{van2017neural, esser2021taming}, we first conduct a toy example as visualized in Figure~\ref{fig:visual_compare}. As shown in the top branch, the image patch corresponding to the single Token \#6302 exhibits an ambiguous visual texture. It is visually difficult to determine whether this local patch belongs to a zebra, a shadow, or random noise. In contrast, the bottom branch displays the result when Token \#6302 is treated as part of a contiguous token phrase (alongside neighbors \#416, \#7747, etc.). Within this coherent sequence, the semantic ambiguity disappeared: the tokens jointly form a clear, continuous zebra stripe pattern. This comparison shows that although Token \#6302 represents the same spatial location, its visual semantic is ambiguous in isolation and becomes recognizable only when integrated into contiguous token phrase. This observation validates a key insight: \textit{visual semantics are not independently stored within individual tokens, but are determined by by the coherent unit with neighboring tokens.}

\begin{figure}[t]
  \centering
  \begin{subfigure}{0.48\linewidth}
    \includegraphics[width=\linewidth]{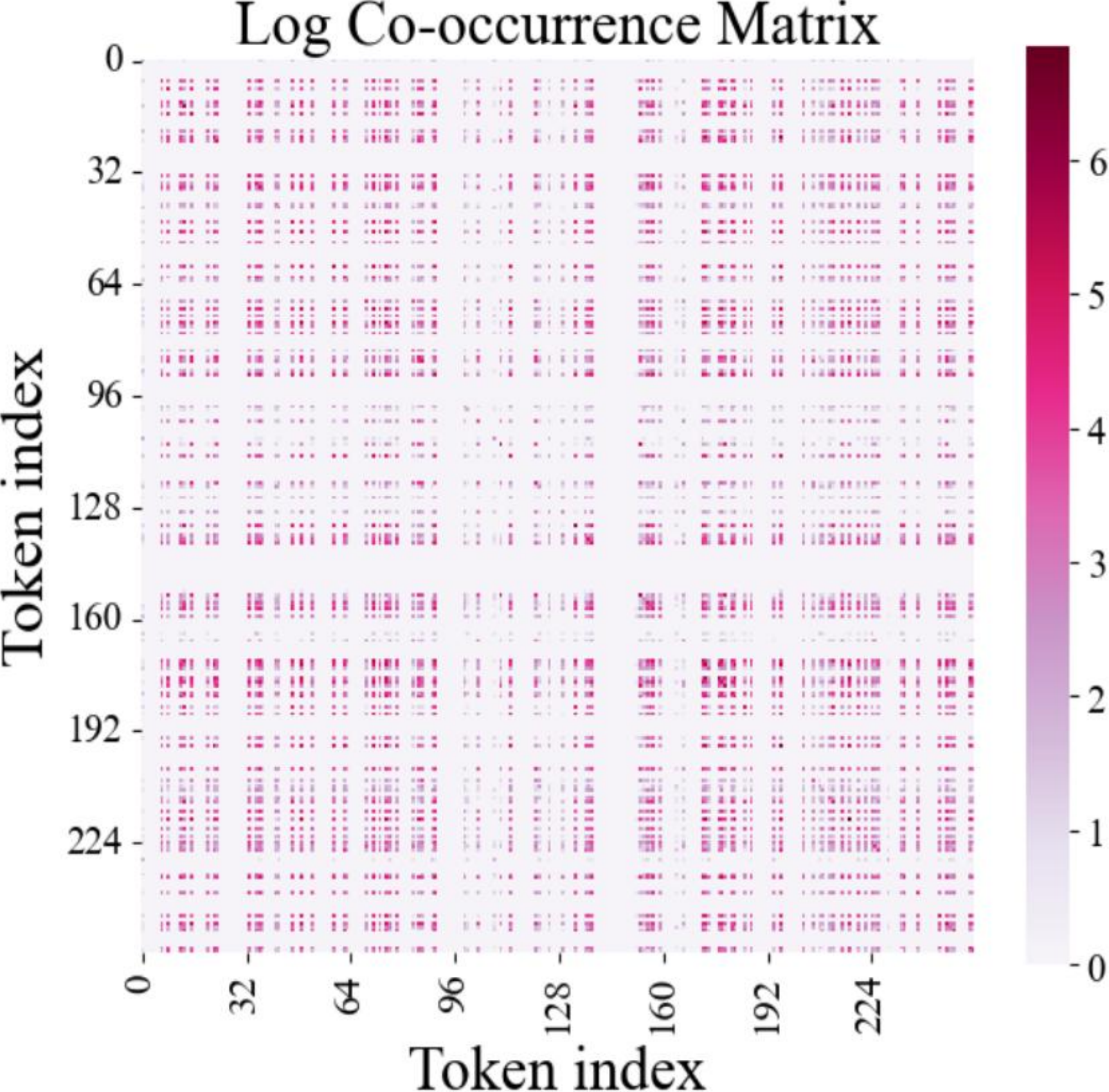}
    \caption{Log co-occurrence of adjacent token pairs.}
    \label{fig:cooc_scatter}
  \end{subfigure}\hfill
  \begin{subfigure}{0.48\linewidth}
    \includegraphics[width=\linewidth]{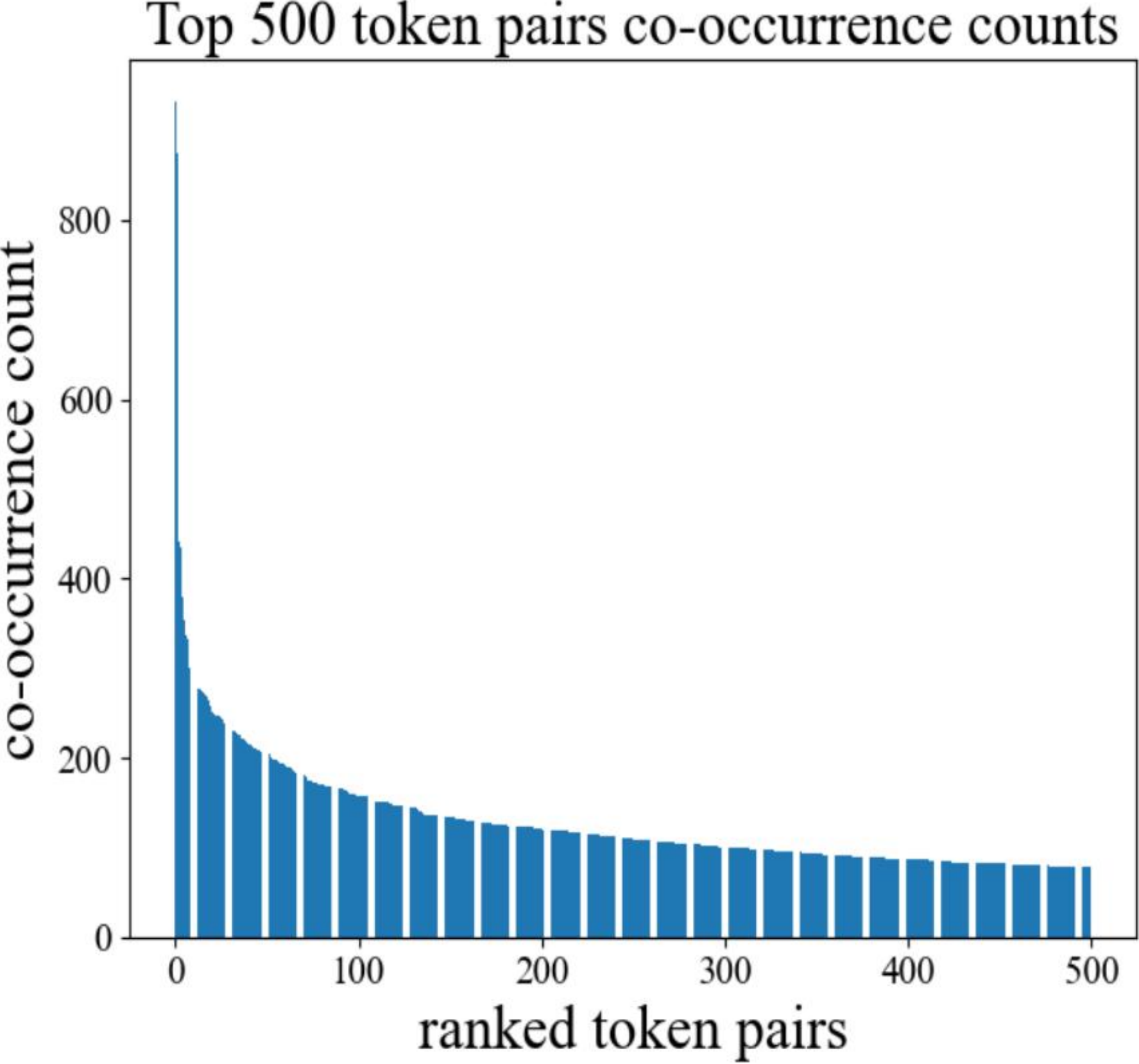}
    \caption{Frequencies of the top-500 adjacent token pairs.}
    \label{fig:cooc_heatmap}
  \end{subfigure}
  \caption{\textbf{Short-range token co-occurrence statistics.} We compute token co-occurrence over consecutive positions in tokenized image sequences, revealing strong local correlations among specific visual tokens.}
  \label{fig:cooc}
\end{figure}

To support this observation with statistical evidence, we analyze the token co-occurrence in large-scale datasets (e.g., MS-COCO 2017) We encode images into discrete token sequences to examine the co-occurrence statistics of adjacent token pairs. As shown in Figure~\ref{fig:cooc}(a), the co-occurrence matrix is distinctively sparse, meaning that adjacent token combinations are constrained to a small subset of patterns. Complementing this, Figure~\ref{fig:cooc}(b) shows that a few frequent pairs account for the vast majority of occurrence. These statistical observations confirm that visual tokens are not independent random variables but are tightly coupled into specific, recurring structures.
This further confirms that \textit{visual semantics are not isolated to an individual token but are inherently encoded as stable, recurring patterns across multiple consecutive tokens}.

\begin{algorithm}[t]
\caption{Phrase Library Construction}
\label{alg:phrase_construction}
\begin{algorithmic}[1]
\REQUIRE Dataset $\mathcal{D}$, Iterations $M$.
\STATE \textbf{Step 1: Iterative Merging}
\STATE Tokenize $\mathcal{D}$ into sequence set $T$.
\FOR{$i = 1$ \textbf{to} $M$}
    \STATE $(u^*, v^*) \leftarrow \arg\max_{(u, v)} \text{Count}(u, v \mid T)$
    \STATE Define new symbol $w \leftarrow (u^*, v^*)$.
    \STATE Update $T$ by replacing all $(u^*, v^*)$ with $w$.
\ENDFOR

\STATE \textbf{Step 2: Recursive Expansion}
\STATE $\mathcal{S} \leftarrow$ Set of all learned symbols $\{w_1, \dots, w_M\}$.
\FOR{\textbf{each} symbol $w \in \mathcal{S}$}
    \STATE Expand $w$ to raw tokens: $p \leftarrow (v_1, v_2, \dots, v_L)$.
\ENDFOR

\STATE \textbf{Step 3: Prefix-based Indexing}
\STATE Initialize $\mathcal{P} \leftarrow \emptyset$.
\FOR{\textbf{each} expanded phrase $p = (v_1, \dots, v_L)$}
    \STATE $\mathcal{P}[v_1] \leftarrow p$ \quad $\triangleright$ Index by starting token
\ENDFOR
\STATE \textbf{return} $\mathcal{P}$
\end{algorithmic}
\end{algorithm}

\subsection{The Proposed SJD-PV}
\label{sec:SJD-PV}


The analysis in Sec.~\ref{sec:motivation} reveals that visual semantics are inherently encoded as coherent token phrases, implying that the speculative verification should align with semantic token phrases rather than isolated tokens. To this end, we propose \textbf{Speculative Jacobi Decoding with Phrase Verification (SJD-PV)}. Different from existing SJD and its variants, we perform the parallel speculative verification step at the token-phrase level rather than on individual tokens. Specifically, we first constructs a token phrase library from large-scale image datasets to capture token phase that represent meaningful semantic. Then, we treat the token phrase library as a prior and design a token-phrase level verification strategy to perform parallel speculative verification. The proposed method is built upon two key steps: \textbf{Phrase Library Construction} and \textbf{Phrase-Level Verification}.

\subsubsection{Phrase Library Construction}
\label{sec:phrase mining}

To capture meaningful visual semantics, we construct a phrase library $\mathcal{P}$ from large-scale image datasets by extracting contiguous token sequences that often appear together. We first encode these raw images into discrete token sequences using a pre-trained tokenizer (e.g., VQ-GAN). On these tokenized sequences, we employ a strategy inspired by Byte Pair Encoding (BPE) to iteratively merge frequently co-occurring token pairs and constitute semantic priors.

The overall construction process is outlined in \textbf{Algorithm~\ref{alg:phrase_construction}}. Formally, we perform $M$ iterations of merging. In each step, we identify the most frequent co-occurring token pair $(u, v)$ in current token sequences, register it as a new merged symbol $w$ (via the rule $(u, v) \rightarrow w$). We then replace all occurrences of this pair with $w$. After $M$ iterations, we obtain a set of merged symbols representing recurring visual semantics. Crucially, since the autoregressive model operates strictly within the original codebook space, these learned high-level symbols must be reverted to verifiable units. We therefore recursively expand each merged symbol $w$ back into its constituent raw token sequence, yielding a concrete phrase $p = (v_1, \dots, v_L)$. Finally, to support efficient $O(1)$ prefix matching during inference, we organize these phrases into a lookup table $\mathcal{P}$, where each phrase is indexed by its starting token $v_1$ (i.e., $\mathcal{P}[v_1] \leftarrow p$).-

\begin{algorithm}[t]
\caption{SJD-PV Verification Strategy}
\label{alg:sjd_pv_verify}
\begin{algorithmic}[1] 
\REQUIRE Draft tokens $\hat{X}_{0:W}$, Verifier $p$, Drafter $q$, Phrase Library $\mathcal{P}$, Vocabulary $\mathcal{V}$, Threshold $\tau$.

\STATE $\triangleright$ \textbf{Step 1: Adaptive Neighborhood Construction}
\FOR {$j = 0$ \textbf{to} $W-1$}
    \STATE $\mathcal{N}_j \leftarrow \{ v \in \mathcal{V} \mid |p(v) - p(\hat{X}_j)| < \tau \}$ 
\ENDFOR

\STATE $\triangleright$ \textbf{Step 2: Speculative Verification Loop}
\STATE $t \leftarrow 0$
\WHILE{$t < W$}
    \STATE $S_{\text{phr}} \leftarrow \mathcal{P}.\text{match}(\hat{X}_t)$
    \STATE $L \leftarrow \text{length}(S_{\text{phr}})$
    \STATE \emph{accepted} $\leftarrow$ \textbf{false}
    
    \IF{$S_{\text{phr}} \neq \text{None}$ \textbf{and} $\forall k, S_{\text{phr}}[k] \in \mathcal{N}_{t+k}$}
        \STATE $\triangleright$ Compute Joint Acceptance Score
        \STATE $\log R_p \leftarrow \sum_{k=0}^{L-1} \left( \log p(S_{\text{phr}}[k]) - \log q(S_{\text{phr}}[k]) \right)$
        
        \STATE $u \sim \mathcal{U}[0, 1]$
        \IF{$\exp(\log R_p) > u$}
            \STATE Append $S_{\text{phr}}$ to $X_{\text{out}}$
            \STATE $t \leftarrow t + L$
            \STATE \emph{accepted} $\leftarrow$ \textbf{true}
        \ENDIF
    \ENDIF
    
    \IF{\textbf{not} \emph{accepted}}
        \STATE $\triangleright$ Fallback: Standard Token Verification
        \STATE Verify single token $\hat{X}_t$ (Standard SJD rule)
        \STATE Append $\hat{X}_t$ to $X_{\text{out}}$ (if accepted)
        \STATE $t \leftarrow t + 1$
    \ENDIF
\ENDWHILE

\STATE \textbf{return} $X_{\text{out}}$

\end{algorithmic}
\end{algorithm}

\subsubsection{Phrase-Level Verification}
\label{sec:SJD-PV-inside-iter}

 The goal of this step is to leverage the semantic prior and verify coherent semantic units rather than isolated tokens. The overall verification procedure is summarized in \textbf{Algorithm~\ref{alg:sjd_pv_verify}}. Formally, for a candidate phrase $p=(v_1, \dots, v_L)$ retrieved from the library $\mathcal{P}$, we evaluate the joint probability ratio between the target model $p$ and the draft model $q$. To ensure numerical stability and avoid underflow when aggregating probabilities over long sequences, we operate in the log-space.  Specifically, the joint acceptance score is calculated as:
\begin{equation}
\log R_p = \sum_{k=1}^{L} \left( \log p(v_k) - \log q(v_k) \right)
\label{eq:phrase_ratio}
\end{equation}
where $v_k$ represents the tokens within the candidate phrase. The candidate phrase is accepted only if:\begin{equation}\exp(\log R_p) > u, \quad \text{where } u \sim \mathcal{U}[0, 1].\label{eq:stochastic_check}\end{equation}If accepted, we commit all $L$ tokens as a whole and advance the window index by $L$ steps.

To apply this verification criterion, we must first match the drafted token sequence with a candidate phrase $p$ from the library $\mathcal{P}$. A straightforward approach would be to require an exact match between the draft and the library entry. However, this strategy often results in a low matching frequency. This occurs because the model output naturally contains uncertainty, meaning multiple different tokens can be equally valid at the same position. Consequently, requiring an exact match discards many semantically valid variations.

To address this, we introduce an \textbf{Adaptive Neighborhood} strategy as shown in Algorithm~\ref{alg:sjd_pv_verify}, Step 1. Drawing inspiration from GSD~\cite{GSD}, which leverages probability proximity to cluster interchangeable tokens, we adapt this concept to enable flexible phrase matching. Instead of a strict one-to-one match, we select if tokens fall within a dynamic set of valid candidates. 
Specifically, for each position $j$ in the Jacobi window, we construct a neighborhood $\mathcal{N}_j$ centered on the drafted token $\hat{X}_j$. We adaptively expand $\mathcal{N}_j$ by including candidate tokens whose probability difference relative to the drafted token is within a threshold $\tau$ (i.e., $|p(v) - p(\boldsymbol{\hat{X}_j})| < \tau$). 
During inference, we scan the Jacobi window and search candidate phrases from $\mathcal{P}$ starting with the current token. A candidate is considered valid only if every constituent token falls within its adaptive neighborhood: 
\[
v_k \in \mathcal{N}_{t+k}, \quad \forall k \in \{0, \dots, L-1\}.
\]
Candidate phrases are then verified using the joint acceptance criterion (Eq.~\ref{eq:phrase_ratio}). If successful, we commit all $L$ tokens simultaneously; otherwise, the method safely falls back to standard token-wise verification.

\subsection{Theoretical Justification for Effectiveness}

\label{subsec:theory_phrase}

To strictly justify the effectiveness of SJD-PV, we analyze the expected acceptance rate directly derived from the rejection sampling mechanism. We demonstrate that verifying a phrase as a joint unit mathematically guarantees a lower bound on the acceptance rate compared to verifying tokens individually, as formally derived below.

\textbf{Definition 1 (Expected Acceptance Rate).}
Let $p$ and $q$ be the target and draft distributions. The expected acceptance rate $\alpha(p, q)$ is defined as the expectation of the acceptance probability ratio:
\begin{equation}
\alpha(p, q) = \mathbb{E}_{x \sim q} \left[ \min\left(1, \frac{p(x)}{q(x)}\right) \right].
\label{eq:alpha_def}
\end{equation}

\textbf{Token-wise Verification.}
In standard SJD, a sequence of $L$ tokens is accepted only if each token is individually accepted. Assuming local independence for analysis, the sequence-level acceptance rate $\alpha_{\text{seq}}$ is the product of individual rates:
\begin{equation}
\alpha_{\text{seq}} = \prod_{i=1}^L \alpha(p_i, q_i) = \prod_{i=1}^L \mathbb{E}_{x_i \sim q_i} \left[ \min\left(1, \frac{p(x_i)}{q(x_i)}\right) \right].
\end{equation}

\textbf{Phrase-level Verification.}
SJD-PV verifies the sequence as a single unit using the joint probability ratio. Let $r_i = \frac{p(x_i)}{q(x_i)}$ be the probability ratio at step $i$. The phrase-level acceptance rate $\alpha_{\text{phr}}$ is governed by the joint ratio $R = \prod_{i=1}^L r_i$:
\begin{equation}
\alpha_{\text{phr}} = \mathbb{E}_{\mathbf{x} \sim q} \left[ \min\left(1, \prod_{i=1}^L r_i \right) \right].
\end{equation}

\textbf{Proposition 1 (Acceptance Rate Improvement).}
For any sequence of draft tokens, the phrase-level acceptance rate is strictly bounded below by the token-wise acceptance rate:
\begin{equation}
\alpha_{\text{phr}} \ge \alpha_{\text{seq}}.
\end{equation}

\textit{Proof.}
We rely on the algebraic inequality of the minimum function. For any non-negative real numbers $a, b \ge 0$, the inequality $\min(1, a \cdot b) \ge \min(1, a) \cdot \min(1, b)$ always~holds.
\begin{itemize}
    \item Case 1 ($a \ge 1, b \ge 1$): $1 \ge 1 \cdot 1$. (Equality holds)
    \item Case 2 ($a < 1, b \ge 1$): The RHS reduces to $a \cdot 1 = a$. For the LHS, since $b \ge 1$, we have $ab \ge a$. Also, by definition $1 > a$. Therefore, $\min(1, ab) \ge a$. Thus, LHS $\ge$ RHS.
    \item Case 3 ($a < 1, b < 1$): $ab = a \cdot b$. (Equality holds)
\end{itemize}
By induction, for a sequence of ratios $r_1, \dots, r_L$:
\begin{equation}
\min\left(1, \prod_{i=1}^L r_i\right) \ge \prod_{i=1}^L \min(1, r_i).
\label{eq:inequality}
\end{equation}
Taking the expectation $\mathbb{E}_{\mathbf{x} \sim q}$ on both sides of Eq.~\eqref{eq:inequality}, and assuming independence such that $\mathbb{E}[\prod (\cdot)] = \prod \mathbb{E}[(\cdot)]$, we obtain $\alpha_{\text{phr}} \ge \alpha_{\text{seq}}$. \hfill $\square$

Mathematically, this result arises because token-wise verification discards the "surplus" confidence of high-probability tokens (where $r_i > 1$) by clipping them to 1. SJD-PV, however, retains the full magnitude of these ratios ($r_i > 1$) to offset the effect of low-confidence tokens ($r_j < 1$). By aggregating probabilities over the sequence, our approach effectively resolves local ambiguity and boosts acceptance.

\section{Experiments}
\begin{table*}[t!]
    \centering
    \small
    \caption{Quantitative evaluation on the MS-COCO 2017 and Parti-prompt.}
    \begin{tabular}{l c c c c c c}
        \toprule
        \textbf{Configuration} & \textbf{Latency (↓)} & \textbf{NFE (↓)} & \multicolumn{2}{c}{\textbf{Acceleration} (↑)} & \textbf{FID (↓)} & \textbf{CLIP-Score (↑)} \\
        \cmidrule(lr){4-5}
        & & & \textbf{Latency} & \textbf{NFE} & & \\
        \midrule
        \multicolumn{7}{l}{\textbf{Parti-prompt}} \\
        Lumina-mGPT \cite{liu2024lumina} & 79.37s & 2392 & 1.00× & 1.00× & -- & 32.091  \\
        Jacobi Decoding \cite{jacobi} & 82.21s & 2300.0 & 0.97x & 1.04× & -- & 32.091 \\
        \midrule
        SJD \cite{SJD} & 36.07s & 1035.3 & 2.15x & 2.31× & -- & 32.090 \\
        \rowcolor{gray!20}
        SJD + Ours & \textbf{34.58s} & \textbf{941.4} & \textbf{2.29x} & \textbf{2.54x} & -- & \textbf{32.098} \\
        \midrule
        GSD (G=3)\cite{GSD} & 33.36s & \textbf 898.97 & 2.38x & 2.66x & -- & 32.11 \\
        \rowcolor{gray!20}
        GSD (G=3) + Ours & \textbf{31.10s} & \textbf{806.8} & \textbf{2.55x} & \textbf{2.96x} & -- & \textbf{32.169} \\
        \midrule
        LANTERN \cite{LANTERN} & 31.40s & \textbf 636.66 & 2.52x & 3.76x & -- & 32.60 \\
        \rowcolor{gray!20}
        LANTERN + Ours & \textbf{29.88s} & \textbf{597.62} & \textbf{2.66x} & \textbf{4.00x} & -- & \textbf{32.71} \\        
        
        \bottomrule
        \midrule
        \multicolumn{7}{l}{\textbf{MS-COCO 2017}} \\
        Lumina-mGPT \cite{liu2024lumina} & 86.55s & 2379 & 1.00× & 1.00× & 30.79 & 31.31  \\
        Jacobi Decoding \cite{jacobi} & 85.64s & 2312 & 1.01x & 1.03x & 30.78 & 31.31 \\
        \midrule
        SJD \cite{SJD} & 40.10s & 1058.6 & 2.22x & 2.25× & 30.78 & 31.31 \\
        \rowcolor{gray!20}
        SJD + Ours&  \textbf{36.47s} & \textbf{945.2}  & \textbf{2.37x} & \textbf{2.52x} & \textbf{30.96} & \textbf{31.40} \\ 
        \midrule
        GSD (G=3)\cite{GSD} & 34.12s & 925.89 & 2.54x & 2.56x & 31.50 & 31.33 \\
        \rowcolor{gray!20}
        GSD (G=3) + Ours& \textbf{32.55s}  & \textbf{803.4} & \textbf{2.66x} & \textbf{2.96x} & 31.11 &  \textbf{31.37} \\ 
        \midrule
        LANTERN \cite{LANTERN} & 33.80s & \textbf 655.37 & 2.35x & 3.65x & 30.72 & 32.7 \\
        \rowcolor{gray!20}
        LANTERN + Ours & \textbf{31.92s} & \textbf{606.28} & \textbf{2.71x} & \textbf{3.92x} & \textbf{30.74} & \textbf{32.72} \\     
        \bottomrule
        \midrule
    \end{tabular}
    \label{tab:main}
    \vspace{-0.3cm}
\end{table*}

\subsection{Experimental Settings}


\noindent \textbf{Datasets and Evaluation Protocol.} 
We evaluate our method on two standard benchmarks: \textbf{Parti-Prompts}~\cite{yu2022scaling} and \textbf{MS-COCO 2017}~\cite{mscoco}. Parti-Prompts consists of 1.6k diverse prompts covering various categories and styles, serving as a standard open-domain benchmark. MS-COCO 2017 is used to test performance under complex, real-world visual semantics. To quantify efficiency, we report \textbf{Latency}, \textbf{NFE} (Number of Function Evaluations), and the \textbf{Acceleration} rate relative to the base model. Furthermore, we assess generation quality using \textbf{FID}~\cite{FID} for visual fidelity and \textbf{CLIP-Score}~\cite{clip} for semantic alignment.

\noindent \textbf{Baselines and Implementation Details.} 
We compare \textbf{SJD-PV} against five representative methods covering distinct decoding paradigms. As the non-accelerated baseline, (1) \textbf{Lumina-mGPT}~\cite{liu2024lumina} serves as the foundational autoregressive backbone. For acceleration comparisons, we include (2) \textbf{Jacobi Decoding}~\cite{jacobi}, which employs parallel iterative refinement to update tokens without speculative verification. Regarding speculative strategies, we compare with (3) \textbf{SJD}~\cite{SJD}, a standard training-free framework that sequentially verifies drafted tokens to reduce inference steps. Furthermore, we evaluate two advanced SJD variants: (4) \textbf{GSD}~\cite{GSD}, which accelerates the process by verifying candidate tokens in clusters, and (5) \textbf{LANTERN}~\cite{jang2024lantern}, which relaxes the acceptance condition by leveraging token interchangeability within the latent space.

All experiments are conducted on a single NVIDIA A100 (80\,GB) GPU using the \emph{same backbone} to ensure fairness. Following GSD~\cite{GSD}, we set the temperature to $1.0$, apply top-$k$ sampling ($k_{\text{img}}=2000, k_{\text{text}}=10$), use a classifier-free guidance scale of $3.0$, and fix the max generation length at $8192$. For Jacobi updates, we follow SJD~\cite{SJD} with a loop interval of $(\ell, r)=(3, (H/16)^2 + H/16 - 10)$, where $H$ is the image height. For GSD, the group size is set to $G=3$. For LANTERN, we set $k = 1$ and $\delta = 0.01$; For our \textbf{SJD-PV}, we construct a phrase library with $8\mathrm{k}$ merges from the MS-COCO training set and employ an adaptive margin $\lambda=0.01$ for neighborhood construction. Since SJD-PV operates purely on verification granularity, it seamlessly falls back to token-wise rules when no phrase matches, ensuring plug-and-play compatibility.
\subsection{Discussion of Quantitative Results}
As shown in Table~\ref{tab:main}, our method significantly accelerates the base Lumina-mGPT across both benchmarks. Specifically, on Parti-Prompts, our best configuration (LANTERN+Ours) reduces decoding latency from 79.37s to 29.88s and NFE from 2392 to 597.62, achieving \textbf{2.66$\times$} latency and \textbf{4.00$\times$} NFE acceleration. Similarly, on MS-COCO 2017, we attain \textbf{2.71$\times$} latency and \textbf{3.92$\times$} NFE speedups. Beyond these gains on base model, our method serves as a universal plug-and-play accelerator, consistently boosting SJD, GSD, and LANTERN (e.g., lifting SJD latency acceleration from $2.22\times$ to $2.37\times$ on MS-COCO 2017). These results confirm that validating coherent phrases effectively mitigates local ambiguity, thereby reducing redundant verification steps.

Besides, our method maintains high generation quality. As shown in Table~\ref{tab:main}, the FID scores on MS-COCO 2017 remain comparable to the baselines (e.g., 30.74 for LANTERN+Ours vs. 30.72 for LANTERN), demonstrating that our phrase-level verification does not compromise the generative fidelity. Notably, we observe a consistent increase in CLIP-Scores across all experiments. For instance, on Parti-Prompts, the CLIP-Score for GSD improves from 32.11 to 32.169 after integrating our method. This suggests that by verifying tokens as coherent phrases, our method preserves the global semantic structure, leading to generated content that aligns more closely with the text prompt.
\begin{figure*}[t]
  \centering
  \includegraphics[width=\textwidth]{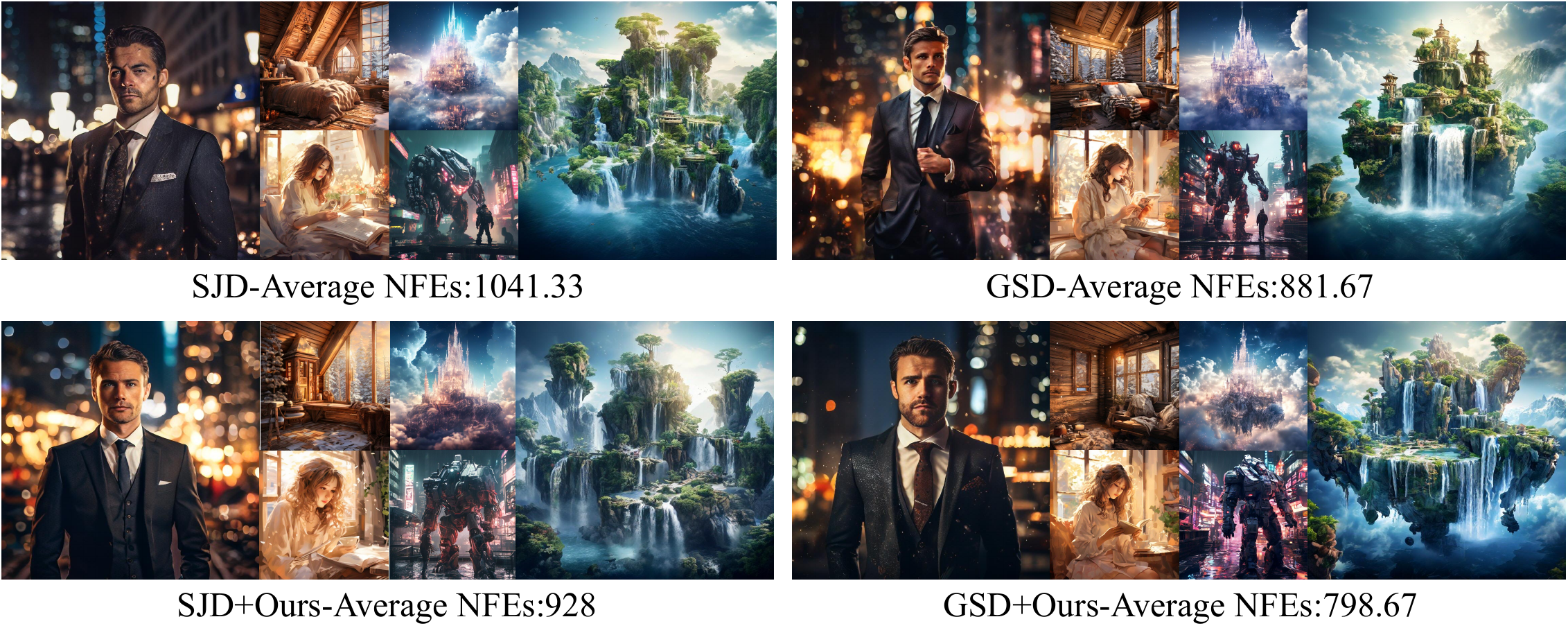}
  \caption{Qualitative experiment. Our method on SJD and GSD shows on significant NFE acceleration while maintaining image quality.}
  \label{fig:qualitative_nfe}
\end{figure*}

\subsection{Discussion of Qualitative Results}
We present qualitative comparisons among SJD~\cite{SJD}, GSD~\cite{GSD}, and their \textbf{SJD-PV}-augmented variants in Figure~\ref{fig:qualitative_nfe}. To comprehensively evaluate visual robustness, we curate prompts covering a broad spectrum of domains: (i) \emph{imaginative and abstract scenes} challenging the synthesis of novel priors; (ii) \emph{photorealistic scenes} requiring dense context and fine-grained textures; (iii) \emph{human portraits}, where structural symmetry and facial geometry are strictly scrutinized; and (iv) \emph{stylized imagery} emphasizing smooth boundaries and coherent artistic textures.

As annotated in Figure~\ref{fig:qualitative_nfe}, our method achieves a visible reduction in computational cost (e.g., reducing average NFEs from 1041.33 to 928 for SJD). Crucially, this acceleration does not compromise visual quality. SJD-PV-augmented decoders preserve high-frequency details (e.g., hair, fabric, foliage) and maintain global spatial coherence. In challenging areas where baseline methods might struggle with local ambiguity, our phrase-level verification ensures structural stability. The output images match the quality of the original model, sometimes delivering even sharper details.This consistent performance across varied domains strongly demonstrates the robust generalization capability of our proposed phrase-level verification strategy.
\subsection{Ablation Study}
\paragraph{Is  Adaptive Neighborhood strategy necessary for phrase matching?}

\begin{table}[t]
\centering
\small
\caption{Ablation of Adaptive Neighborhood strategy}
\label{tab:ablation_neighborhood}

\renewcommand\tabcolsep{10pt}

\begin{tabular}{l|ccc}
\toprule
\textbf{Setting} & \textbf{Latency} $\downarrow$ & \textbf{NFE} $\downarrow$ & \textbf{CLIP} $\uparrow$ \\
\midrule
w/o & 35.96s & 998.3 & 32.090 \\
\rowcolor{gray!15} \textbf{SJD-PV (Ours)} & \textbf{34.58s} & \textbf{941.4} & \textbf{32.098} \\
\bottomrule
\end{tabular}
\end{table}
We conduct an ablation on the Parti-Prompts dataset by comparing our full method against a baseline using strict exact matching (denoted as "w/o" in Table 2). As shown in the table, removing the Adaptive Neighborhood strategy notably increases computational cost. In contrast, our adaptive approach secures efficiency gains while maintaining a stable CLIP score. This confirms that probability-based relaxation is essential for capturing semantically valid phrases that strict matching would otherwise discard.

\begin{table}[t]
\centering
\small
\caption{Ablation of iterations $M$}
\label{tab:ablation_iterations}

\renewcommand\tabcolsep{10pt}

\begin{tabular}{l|ccc}
\toprule
\textbf{Setting} & \textbf{Latency} $\downarrow$ & \textbf{NFE} $\downarrow$ & \textbf{CLIP} $\uparrow$ \\
\midrule
M=4k  & 37.02s & 1005.6 & 32.072 \\
\rowcolor{gray!15} \textbf{M=8k} & \textbf{34.58}s & \textbf{941.4} & \textbf{32.098}  \\
M=16k  & 34.60s & 942.1 & 31.790 \\
\bottomrule
\end{tabular}
\end{table}
\paragraph{How does the number of merging iterations affect performance?}
The number of merging iterations $M$ controls the length of the learned phrases. Increasing $M$ produces longer phrases that can decode more tokens at once, but they also become more specific and harder to match. To analyze this trade-off, we conduct an ablation on the Parti-Prompts dataset (Table 3). As shown, increasing $M$ from $4\text{k}$ to $8\text{k}$ yields clear benefits, significantly reducing NFE while improving the CLIP score . However, further increasing $M$ to $16\text{k}$ offers no additional efficiency gain and notably degrades visual fidelity . This suggests that overly long phrases suffer from data sparsity, making $M=8\text{k}$ the optimal setting for our phrase library.

\paragraph{What value should neighborhood threshold $\tau$ take?}
The threshold $\tau$ controls the strictness of phrase matching. As shown in Table 3, a conservative $\tau=0.005$ limits acceleration (NFE 965.2). Relaxing it to $\tau=0.01$ significantly boosts efficiency (NFE 941.4) while maintaining optimal image quality. However, an overly loose threshold ($\tau=0.05$) introduces noise, causing the CLIP score to drop. Thus, we select $\tau=0.01$ as the optimal setting.

\begin{table}[t!]
\small
\caption{
Ablation study on the neighborhood threshold $\tau$.
}
\renewcommand\tabcolsep{8pt}
\centering
\smallskip
\scalebox{1.0}{
\begin{tabular}{lcccc} 
\toprule
\textbf{Threshold} ($\tau$) & \textbf{Latency} & \textbf{NFE} $\downarrow$ & \textbf{CLIP-Score} $\uparrow$ \\
\midrule   
$\tau=0.005$ & 35.12s & 965.2 & 32.095 \\
\rowcolor{gray!15} \textbf{$\tau=0.01$} & \textbf{34.58s} & \textbf{941.4} & \textbf{32.098} \\
$\tau=0.02$ & 34.20s & 928.5 & 32.012 \\
$\tau=0.05$ & 33.85s & 915.0 & 31.650 \\
\bottomrule
\end{tabular}
}
\label{tab:ablation_tau}
\end{table}


\section{Conclusion}
We introduce \textbf{SJD-PV}, a training-free framework that shifts speculative verification from the token level to the phrase level. By validating coherent semantic units rather than isolated tokens, our approach resolves the local ambiguity in autoregressive generation. Crucially, SJD-PV serves as a plug-and-play module. It integrates seamlessly with existing SJD methods without retraining. Extensive experiments confirm that our method significantly accelerates inference, achieving a superior efficiency-quality trade-off compared to state-of-the-art approaches.

\nocite{langley00}

\bibliography{example_paper}
\bibliographystyle{icml2026}




\end{document}